\documentclass{article} % For LaTeX2e
\usepackage[preprint]{colm2026_conference}

\usepackage{microtype}
\usepackage{hyperref}
\usepackage{url}
\usepackage{graphicx}
\usepackage{amsmath}
\usepackage{amssymb}
\usepackage{mathtools}
\usepackage{amsthm}
\usepackage{enumitem}

\usepackage[table,dvipsnames]{xcolor}

\definecolor{HL}{RGB}{230,232,246} % light lavender for highlight rows
\definecolor{Gain}{RGB}{26,140,26} % green
\definecolor{Loss}{RGB}{180,20,20} % red
\definecolor{DeepGreen}{RGB}{0,128,0}
\definecolor{DeepRed}{RGB}{180,0,0}

\usepackage{booktabs,multirow,subcaption}
\usepackage[most]{tcolorbox}

\usepackage{lineno}

\definecolor{darkblue}{rgb}{0, 0, 0.5}
\hypersetup{colorlinks=true, citecolor=darkblue, linkcolor=darkblue, urlcolor=darkblue}

\theoremstyle{plain}

\theoremstyle{definition}

\theoremstyle{remark}

\title{Start Classifying: Categorical Critics for LLM Reinforcement Learning}

\author{%
\small
\textbf{Zhijian Zhou}$^{\ast,1,2,3}$ \quad
\textbf{Long Li}$^{\ast,4}$ \quad
\textbf{Xuan Zhang}$^{1,2,3}$ \quad
\textbf{Zongkai Liu}$^{2}$ \quad
\textbf{Yulei Qin}$^{3}$ \\
\textbf{Ke Li}$^{3}$ \quad
\textbf{Xing Sun}$^{3}$ \quad
\textbf{Xiaoyu Tan}$^{3,\dagger,\ddagger}$ \quad
\textbf{Chao Qu}$^{1,5,\dagger}$ \quad
\textbf{Yuan Qi}$^{1,5,\dagger}$ \\[3pt]
\normalfont
$^{1}$Fudan University \quad
$^{2}$Shanghai Innovation Institute \quad
$^{3}$Tencent YoutuLab \\
$^{4}$Griffith University \quad
$^{5}$Shanghai Academy of Artificial Intelligence for Science \\
\texttt{arthurtan@tencent.com} \quad
\texttt{quchao@fudan.edu.cn} \quad
\texttt{qiyuan@fudan.edu.cn} \\
$^{\ast}$Equal contribution \quad
$^{\dagger}$Corresponding authors \quad
$^{\ddagger}$Project leader
}

\begin{document}

\ifcolmsubmission
\linenumbers
\fi

\maketitle

\begin{abstract}
Proximal Policy Optimization (PPO) for large language models typically trains its critic by mean-squared-error (MSE) regression on scalar value targets. Although scalar MSE is statistically valid for estimating the conditional expected return, sparse binary rewards in reinforcement learning with verifiable rewards (RLVR) make critic optimization and calibration especially consequential: small value errors directly distort the scalar advantages used by PPO. We study whether a classification-based training objective can improve this critic signal. \textbf{HL-Gauss PPO} replaces the scalar MSE head with a categorical predictor over a discretized value support, trained by cross-entropy against smoothed HL-Gauss targets. Its output is decoded to a scalar expectation for standard GAE and PPO; the actor update is therefore unchanged and is not distributional. Across mathematical reasoning, tool-augmented math, and Search-R1, and on both Qwen2.5 and Qwen3 backbones, HL-Gauss PPO consistently improves over strong PPO and DAPO baselines. Controls with one-hot, two-hot, and Bernoulli two-bin critics show that neither a larger output head nor binary classification alone explains the gains. On a common collection of reasoning prefixes, HL-Gauss improves Brier score and calibration error and yields more symmetric, lower-variance advantages. These results position categorical value learning as an effective optimization surrogate for PPO critics in RLVR.
\end{abstract}

\section{Introduction}
\label{sec:introduction}

Reinforcement learning has emerged as a powerful paradigm for improving the reasoning capabilities of large language models (LLMs). Recent work on reinforcement learning with verifiable rewards (RLVR) --- where a deterministic verifier provides binary correctness signals --- has demonstrated striking gains on mathematical reasoning~\citep{jaech2024openai,deepseek2025r1,yu2025dapo,shao2024deepseekmath}. Within this paradigm, two broad families of algorithms have gained traction: critic-free methods such as GRPO~\citep{shao2024deepseekmath} and DAPO~\citep{yu2025dapo}, which estimate advantages from within-group reward comparisons; and actor--critic methods based on PPO~\citep{schulman2017ppo}, which learn an explicit value function to compute advantages via GAE~\citep{schulman2015gae}. While critic-free methods sidestep the difficulty of value learning, recent studies such as VAPO~\citep{zhong2025vapo} have shown that a well-trained critic can provide stable and sample-efficient training for long-horizon reasoning tasks.

Despite the renewed interest in PPO for LLMs, comparatively little attention has been paid to the critic's \emph{learning objective}. Standard PPO fits scalar value targets with MSE. In the common RLVR setting with binary terminal rewards and $\gamma{=}\lambda{=}1$, the return from an intermediate state is Bernoulli and the population-optimal scalar prediction is its conditional mean, equivalently the probability of eventual success. Scalar MSE is therefore statistically consistent for the value required by PPO; binary or bimodal returns do not make scalar value learning invalid. The practical issue is optimization and calibration. Because the critic enters the actor update through $\hat A_t=G_t-V_\phi(s_t)$, modest value miscalibration can create substantially asymmetric positive and negative advantages, as we show in Section~\ref{sec:advantage_quality}.

Recasting regression as classification is well established: histogram losses, categorical value heads, and broader empirical studies show optimization benefits in conventional RL~\citep{imani2018improving,schrittwieser2020mastering,farebrother2024stop}. DisPPO instead models the full return distribution in LLM PPO via quantile regression~\citep{zhoudisppo}. We use classification only as a critic-training surrogate and decode a scalar expectation before an unchanged GAE/PPO update; our method is therefore not a distributional policy-improvement algorithm. This isolates whether a smoothed categorical critic objective can improve calibration and advantage estimation without changing policy optimization.

We propose \textbf{HL-Gauss PPO}, which replaces the scalar MSE critic head with a categorical value head trained via HL-Gauss cross-entropy~\citep{imani2018improving}, while leaving the actor-side PPO update entirely unchanged. This is a minimal, drop-in modification: only the critic's output layer ($\mathbb{R}^d\!\to\!\mathbb{R}^1$ becomes $\mathbb{R}^d\!\to\!\mathbb{R}^m$) and critic loss change. Empirically, HL-Gauss PPO consistently outperforms strong PPO and DAPO baselines across mathematical reasoning, tool-augmented math, and Search-R1. The trend also holds on Qwen3-4B-Base, extending the main Qwen2.5-7B evaluation.

We study two possible optimization differences---the per-sample gradient channel induced by the output head and the local geometry induced by target smoothing---without treating either as a complete causal explanation. More directly, calibration measurements and advantage dynamics show that the MSE critic becomes overconfident on likely-failure prefixes and produces strongly asymmetric raw advantages, whereas HL-Gauss yields better calibrated scalar predictions and more balanced signals. A Bernoulli two-bin baseline and matched categorical-head ablations further separate binary classification, head capacity, and target smoothing.

\begin{figure}[t]
\centering
\includegraphics[width=\linewidth]{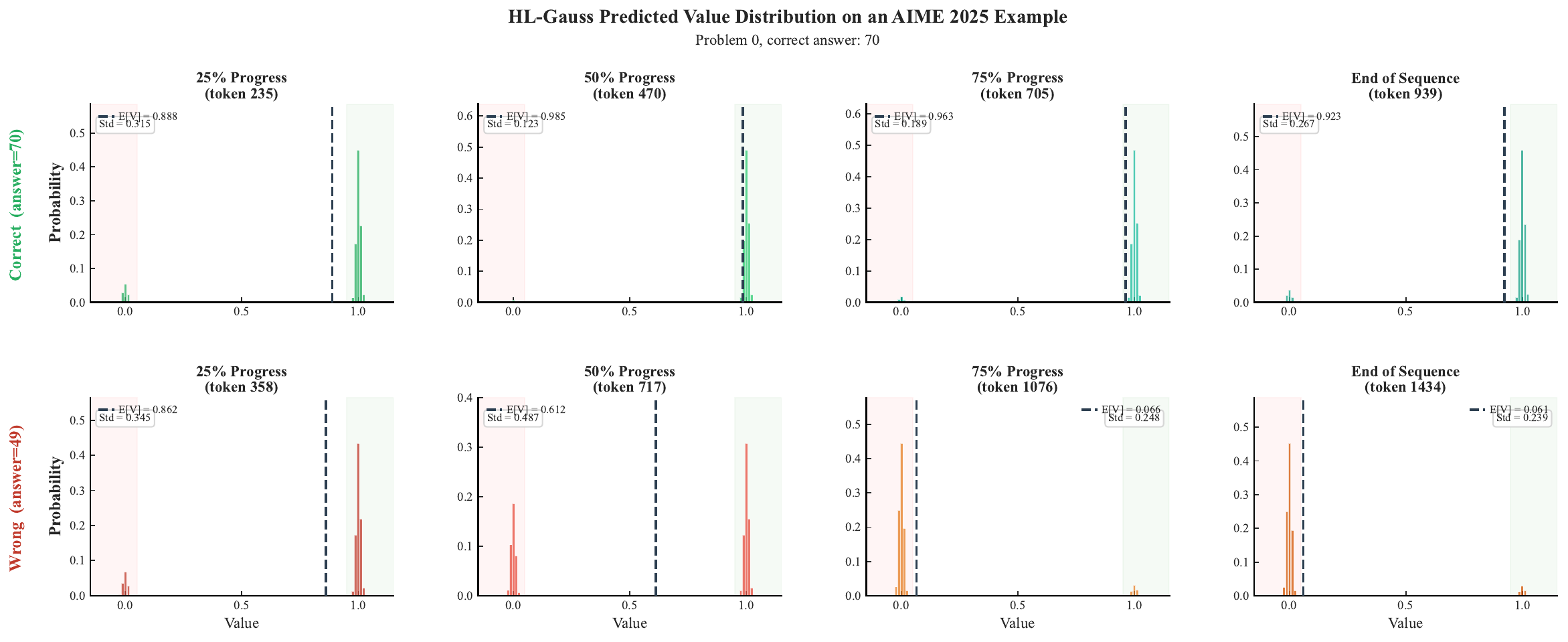}
\caption{Categorical critic outputs across problem-solving progress on a paired AIME 2025 example. Early prefixes have diffuse mass near both endpoints; the output subsequently concentrates near 1 for the correct rollout and near 0 for the incorrect rollout. PPO uses only the decoded expectation of each distribution, so this visualization describes the critic parameterization rather than a distributional actor update.}
\label{fig:hl_gauss_value_dist}
\end{figure}

Our contributions are as follows:
\begin{itemize}[nosep,leftmargin=1.5em]
\item \textbf{Controlled evaluation in LLM RLVR.} We apply the established HL-Gauss value-learning objective to PPO critics for LLM reasoning and show consistent gains across mathematical reasoning, tool use, search, and Qwen2.5/Qwen3 backbones while leaving the actor update unchanged.
\item \textbf{Controls and diagnostics.} One-hot, two-hot, Bernoulli two-bin, support, and smoothing controls isolate the role of target construction. Direct calibration metrics and advantage analyses connect the critic objective to the scalar learning signal received by PPO.
\end{itemize}

\section{Preliminaries}
\label{sec:preliminaries}

This section reviews the two foundations underlying our method: PPO-style reinforcement learning for large language models, together with several practical refinements used in recent reasoning-RL systems, and the regression-as-classification perspective for value learning.

\subsection{PPO-style RL for LLMs and practical refinements}
\label{sec:prelim-ppo}

We model autoregressive generation as a sequential decision process. Let $x$ denote the input prompt and $y_{1:T}$ the generated token sequence. At decoding step $t$, the state is the prefix $s_t=(x, y_{1:t-1})$ and the action is the next token $a_t = y_t$. The policy $\pi_\theta(a_t \mid s_t)$ is parameterized by the LLM with parameters $\theta$. The learning objective is to maximize expected return $J(\theta) = \mathbb{E}_{\pi_\theta}\!\left[\sum_{t=1}^{T} \gamma^{t-1} r_t \right]$, where rewards may be provided at token or trajectory level.

Standard PPO uses a clipped surrogate objective optimized via policy gradients, with advantages estimated using generalized advantage estimation (GAE)~\citep{schulman2015gae,schulman2017ppo}: $\hat{A}_t = \sum_{l=0}^{\infty}(\gamma\lambda)^l \delta_{t+l}$ where $\delta_t = r_t + \gamma V_\phi(s_{t+1}) - V_\phi(s_t)$. The critic $V_\phi(s)$ predicts a scalar value estimate and is typically trained by MSE regression to a bootstrapped return target.

In modern reasoning-RL systems for LLMs, this basic PPO recipe is often augmented with practical refinements (e.g., asymmetric clipping, token-level policy gradients, group sampling, value pretraining)~\citep{zhong2025vapo} motivated by long-horizon reasoning and sparse rewards. Our focus is on changing how the critic represents and learns value targets, not on the actor objective or training recipe.

\subsection{Regression as classification }
\label{sec:prelim-hlgauss}
\cite{farebrother2024stop} propose to train value functions as a classification problem. The key idea is to project a scalar target $y$ onto a fixed discrete support $\{z_i\}_{i=1}^m$, predict a categorical distribution $\mathbf{p}_\phi(s)$ via an $m$-way softmax head, and optimize a cross-entropy loss instead of an MSE regression loss. Given a value range $[v_{\min}, v_{\max}]$, the interval is discretized into $m$ bins with centers $\{z_i\}$. The value prediction is the expected value of the categorical distribution:
\[
\hat{V}_\phi(s)=\sum_{i=1}^m p_\phi(z_i\mid s)\, z_i .
\]

The target distribution $\mathbf{q}(y)$ can be constructed via sharp (one-hot) or smooth (Gaussian) encodings. \textbf{HL-Gauss} \citep{imani2018improving} applies a Gaussian kernel:
\[
q_i(y)\propto \exp\left(-\frac{(z_i-y)^2}{2\sigma^2}\right), \quad \sum_i q_i(y)=1 .
\]
The critic is trained with cross-entropy loss: $L_V(\phi)=-\sum_i q_i(y)\log p_\phi(z_i\mid s)$.

\section{Method: HL-Gauss PPO}
\label{sec:method}

We propose HL-Gauss PPO, which replaces the scalar regression critic in PPO with a categorical value head trained via the HL-Gauss cross-entropy objective. The actor-side PPO update, advantage estimation, and rollout pipeline remain unchanged; only the critic architecture and its learning target differ. The method is therefore a classification-based surrogate for learning the scalar value consumed by PPO, not a distributional policy-improvement algorithm. Architectural and hyperparameter details are deferred to Appendix~\ref{app:method-impl}.

\subsection{Categorical value head}
\label{sec:method-head}

In PPO, the critic appends a linear projection $\mathbf{w}^\top \mathbf{h} + b$ to the backbone's last hidden state $\mathbf{h}\in\mathbb{R}^d$, producing a scalar value estimate $V_\phi(s)\in\mathbb{R}$. We replace this with an $m$-dimensional linear layer followed by a softmax, yielding a categorical distribution over a fixed discrete support.

Concretely, we partition the value range $[v_{\min}, v_{\max}]$ into $m$ equally spaced bins with width $\Delta=(v_{\max}-v_{\min})/m$ and centers $z_i = v_{\min} + (i - \tfrac{1}{2})\,\Delta$ for $i=1,\dots,m$. The critic outputs logits $\boldsymbol{\ell}(s) = W\mathbf{h}+\mathbf{b} \in \mathbb{R}^m$, and the predicted distribution is $p_i(s)=\mathrm{softmax}(\boldsymbol{\ell}(s))_i$. The scalar value estimate is recovered as the expectation:
$\hat{V}_\phi(s) = \sum_{i=1}^{m} p_i(s)\, z_i.$ This scalar value is used downstream for GAE advantage computation (Section~\ref{sec:prelim-ppo}); no higher-order distributional statistic is passed to the actor.

\subsection{Target distribution construction}
\label{sec:method-target}

At each training step, the critic receives a scalar bootstrapped return $y_t = \hat{V}_\phi(s_t) + \hat{A}_t$ computed via GAE. We project $y_t$ onto a target distribution $\mathbf{q}(y_t)$ over the $m$ bins. For the HL-Gauss projection, each bin's target probability is computed as the Gaussian CDF mass falling within that bin:
\[
q_i(y_t) \propto \Phi\!\left(\frac{z_i + \Delta/2 - y_t}{\sigma}\right) - \Phi\!\left(\frac{z_i - \Delta/2 - y_t}{\sigma}\right),
\]
where $\Phi(\cdot)$ is the standard normal CDF and $\sigma$ controls the smoothness of the target \citep{imani2018improving,farebrother2024stop}. The resulting distribution is normalized to sum to one.

\subsection{Training objective}
\label{sec:method-objective}

The critic is trained by minimizing the cross-entropy between the target distribution and the predicted distribution:
\[
L_{\mathrm{critic}}(\phi) = -\frac{1}{|\mathcal{M}|}\sum_{t\in\mathcal{M}} \sum_{i=1}^{m} q_i(y_t)\,\log p_i(s_t),
\]
where $\mathcal{M}$ is the set of valid (non-padded) response tokens. The actor objective remains the standard PPO clipped surrogate $L_{\mathrm{clip}}(\theta)$ . Actor and critic are updated alternately within each PPO epoch, following standard practice.

\section{Analysis: Optimization and advantage calibration}
\label{sec:theory}

In binary-reward RLVR, scalar MSE is population-consistent for the conditional expected return. Our analysis therefore does not argue that scalar critics are representationally invalid. Instead, we describe two local optimization differences between scalar and categorical heads and then connect practical critic calibration to the advantages used by PPO.

\subsection{Per-sample gradient channels}
Standard PPO trains a scalar critic $V_\phi(s) = \mathbf{w}^\top \mathbf{h} + b$, where $\mathbf{h} \in \mathbb{R}^d$ is the last-layer hidden state of the LLM backbone. Minimizing the MSE loss $\mathcal{L}_{\text{MSE}} = (V_\phi(s) - y)^2$ yields a gradient with respect to the backbone representation:
\begin{equation}
    \nabla_{\mathbf{h}} \mathcal{L}_{\text{MSE}} = 2(V_\phi(s) - y) \cdot \mathbf{w} \in \text{span}(\mathbf{w}).
\end{equation}
For a fixed scalar head and one sample, this gradient is confined to the direction $\mathbf w$. This is a statement about the local gradient passed to the final hidden state; it does \emph{not} imply that cumulative backbone updates across samples, steps, or layers are globally rank one.

In contrast, a categorical critic with $m$ bins uses logits $\mathbf{z} = W\mathbf{h} + \mathbf{b}$ with $W \in \mathbb{R}^{m \times d}$ and $\mathbf{p}=\mathrm{softmax}(\mathbf{z})$. Given target $\mathbf{q}$, cross-entropy yields $\nabla_{\mathbf{h}} \mathcal{L}_{\text{CE}} = W^\top (\mathbf{p} - \mathbf{q})$, which can combine multiple rows of $W$. The categorical head therefore provides a broader \emph{per-sample} gradient channel. This observation is generic to multi-output heads and is not, by itself, an explanation specific to RLVR. Indeed, the weaker one-hot results in Table~\ref{tab:avg-k-results} show that a multi-output head alone is insufficient.

\subsection{Target smoothing and local geometry}
The difference between one-hot and HL-Gauss controls points to target construction rather than output dimensionality alone. Sharp one-hot targets encourage predictions near simplex corners, whereas HL-Gauss assigns probability mass to neighboring bins and encourages an interior solution.

HL-Gauss mitigates this by applying Gaussian smoothing to the value targets. From an information-geometric perspective, HL-Gauss keeps the target distribution away from the corners of the probability simplex. For a categorical head with logits $\boldsymbol{\ell}$ and probabilities $\mathbf{p}=\mathrm{softmax}(\boldsymbol{\ell})$, the Fisher information matrix with respect to the logits is
\begin{equation}
    F(\boldsymbol{\ell})
    =
    \mathbb{E}_{y \sim \mathbf{p}}
    \left[
    \nabla_{\boldsymbol{\ell}} \log p(y;\boldsymbol{\ell})
    \nabla_{\boldsymbol{\ell}} \log p(y;\boldsymbol{\ell})^\top
    \right]
    =
    \mathrm{diag}(\mathbf{p}) - \mathbf{p}\mathbf{p}^\top .
\end{equation}
When $\mathbf{p}$ approaches a corner of the simplex, this matrix becomes nearly degenerate. Near the population optimum, smoothed targets encourage less extreme predictions and can yield better-conditioned local geometry. We treat this as a plausible optimization mechanism, supported by the one-hot, two-hot, and bandwidth controls, rather than as a complete causal account of the policy gains.

\subsection{Calibration and advantage asymmetry}

Critic calibration directly shapes the PPO update. In the illustrative case $\gamma{=}\lambda{=}1$, GAE reduces to $\hat{A}_t=G_t-V_\phi(s_t)$ with $G_t\in\{0,1\}$. If the scalar prediction overestimates the conditional success probability, failed trajectories receive negative advantages of magnitude $V_\phi(s_t)$ while successful trajectories receive positive advantages of magnitude $1-V_\phi(s_t)$. Thus even a valid scalar estimator can produce asymmetric learning signals when imperfectly optimized or calibrated under a changing on-policy distribution. Our empirical question is whether the HL-Gauss training surrogate reduces this error in practice. Section~\ref{sec:advantage_quality} answers affirmatively with direct Brier/ECE measurements and advantage statistics. The categorical output itself is not passed to the actor; the benefit appears through the quality of its decoded scalar value.

\section{Experiments}
\label{sec:experiments}

We evaluate HL-Gauss PPO to study whether a classification-based critic objective improves reinforcement learning for LLMs. Our central comparison is against PPO with a standard scalar MSE critic, which isolates the effect of replacing scalar value regression with categorical value prediction. We additionally compare against DAPO~\citep{yu2025dapo} and conduct \emph{ablations} over a Bernoulli two-bin critic and one-hot, two-hot, and HL-Gauss targets.

Our experimental design is intentionally controlled: unless otherwise stated, all methods share the same actor objective, rollout and sampling pipeline, optimization schedule, and evaluation protocol, and differ only in how the critic target is parameterized and trained. This setup attributes differences to the critic objective rather than to changes in the policy optimization recipe. We additionally report tool-augmented math (Appendix~\ref{app:math-with-tool}), Qwen3 generalization (Appendix~\ref{app:qwen3-generalization}), smoothing-bandwidth ablations (Appendix~\ref{app:ablation-sigma}), and computational overhead (Appendix~\ref{app:compute}).

\subsection{Experimental setup}
\label{sec:exp_setup}

Unless otherwise stated, all methods are trained under a unified PPO-style framework, where the only change across variants is the critic learning target: standard PPO uses scalar regression with an MSE loss, while our categorical variants use one-hot, two-hot, or HL-Gauss targets. In particular, the actor-side PPO loss, rollout generation procedure, sampling budget, optimizer settings, learning-rate schedule, and evaluation scripts are kept fixed across all methods.

To make the comparison meaningful, we adopt a strong PPO baseline rather than a minimal vanilla implementation. Specifically, our baseline incorporates several optimization techniques that have become common in recent large-scale RL for reasoning models, following \emph{DAPO}~\citep{yu2025dapo} and parts of the VAPO training recipe~\citep{zhong2025vapo}, including Clip-Higher, Token-level Loss, Critic Pretraining, and Group Sampling. These components are applied uniformly across all critic variants.

We nevertheless refer to this baseline as \emph{PPO} rather than \emph{VAPO}, since we do not adopt the full VAPO formulation. In particular, we exclude Length-Adaptive GAE and Positive Example Loss. Empirically, we find that using $\lambda = 1$ in actor performs best in our setting, which leaves little room for additional gains from Length-Adaptive GAE. We also find that Positive Example Loss has a disproportionate effect on pass@$k$, making it less suitable for our main controlled comparison, whose goal is to isolate the effect of the critic parameterization.

\textbf{Model setup.} We use Qwen2.5-Math-7B~\citep{yang2024qwen2} as the backbone for mathematical reasoning and Qwen2.5-7B-Instruct~\citep{qwen2025qwen25technicalreport} for tool-augmented math and Search-R1 experiments. We additionally test Qwen3-4B-Base~\citep{qwen3technicalreport} under the same PPO protocol (Appendix~\ref{app:qwen3-generalization}). For each backbone, PPO and HL-Gauss PPO start from the same initialization and use the same tokenizer, context setting, and generation/evaluation pipeline. All experiments are implemented in veRL, an open-source RLHF framework built on HybridFlow~\citep{shao2024verl}.

\textbf{Dataset setup.} For training on mathematical reasoning, we use DAPO-Math-17K~\citep{yu2025dapo}. We evaluate on AIME24, AIME25 ~\citep{aime}, and further report HMMT~\citep{hmmt}, BeyondAIME, and Brumo~\citep{yu2025dapo} for broader coverage. For an agentic task, Search-R1, we report mean@1 accuracy on NQ~\citep{kwiatkowski2019natural}, TriviaQA~\citep{joshi2017triviaqa}, PopQA~\citep{mallen2023popqa}, HotpotQA~\citep{yang2018hotpotqa}, 2Wiki~\citep{ho2020constructing}, MuSiQue~\citep{trivedi2022musique}, and Bamboogle~\citep{press2023measuring}. Across all tables, we report avg@\emph{k} and pass@\emph{k} where applicable, and use $\Delta$ to denote absolute gains over PPO under matched settings.

\subsection{Mathematical reasoning}
\label{sec:math_results}

Table~\ref{tab:avg-k-results} shows that HL-Gauss PPO consistently outperforms PPO and DAPO under the same experimental setup. On Qwen2.5-Math-7B, HL-Gauss PPO improves the overall avg@256 score from 15.88 (PPO) to 18.74 (+2.86), with the largest single-dataset gain on AIME25 (+5.20). The same trend appears in pass@256, where the average score increases from 38.48 to 48.06 (+9.58). On Qwen3-4B-Base, the corresponding averages improve from 15.33 to 17.18 for avg@256 and from 53.40 to 57.13 for pass@256 (Appendix~\ref{app:qwen3-generalization}), showing that the trend is not confined to Qwen2.5-7B.
\begin{table*}[t]
\centering
\caption{Performance on mathematical reasoning with avg@\emph{k}. Best results in \textbf{bold}, second best \underline{underlined}, within each backbone group. $\Delta$ rows denote the performance gain of HL-Gauss PPO over PPO.}
\label{tab:avg-k-results}

\begin{subtable}{\textwidth}
\centering
\caption{avg@\emph{k} results}
\setlength{\tabcolsep}{4pt}
\renewcommand{\arraystretch}{1.1}
\resizebox{\textwidth}{!}{%
\begin{tabular}{lcccccc}
\toprule
& \textbf{AIME24} & \textbf{AIME25} & \textbf{HMMT} & \textbf{BeyondAIME} & \textbf{Brumo} & \textbf{Avg.} \\
& \textit{avg@256} & \textit{avg@256} & \textit{avg@256} & \textit{avg@256} & \textit{avg@256} & \\
\midrule
\emph{Qwen2.5-Math-7B}   & 10.70 & 4.40 & 0.30 & 2.00 & 5.72 & 4.62 \\
\quad + DAPO             & 31.90 & 17.30 & \underline{4.28} & 7.50 & 19.00 & 16.00 \\
\quad + PPO              & 34.30 & 18.60 & 1.00 & 7.30 & 18.20 & 15.88 \\
\quad + Bernoulli 2-bin PPO & 34.10 & 17.40 & 3.00 & 7.50 & 15.50 & 15.50 \\
\quad + One-hot PPO      & 27.80 & 17.40 & 1.50 & \underline{7.80} & 19.80 & 14.86 \\
\quad + Two-hot PPO      & \textbf{37.40} & \underline{19.20} & 3.23 & 7.60 & \underline{20.30} & \underline{17.55} \\
\rowcolor{blue!8}
\quad + HL-Gauss PPO (ours) & \underline{35.20} & \textbf{23.80} & \textbf{4.78} & \textbf{8.50} & \textbf{21.40} & \textbf{18.74} \\
\quad $\Delta$ &
\textcolor{ForestGreen}{+0.90} &
\textcolor{ForestGreen}{+5.20} &
\textcolor{ForestGreen}{+3.78} &
\textcolor{ForestGreen}{+1.20} &
\textcolor{ForestGreen}{+3.20} &
\textcolor{ForestGreen}{+2.86} \\
\bottomrule
\end{tabular}
}
\end{subtable}

\vspace{0.8em}

\begin{subtable}{\textwidth}
\centering
\caption{pass@\emph{k} results}
\setlength{\tabcolsep}{4pt}
\renewcommand{\arraystretch}{1.1}
\resizebox{\textwidth}{!}{%
\begin{tabular}{lcccccc}
\toprule
& \textbf{AIME24} & \textbf{AIME25} & \textbf{HMMT} & \textbf{BeyondAIME} & \textbf{Brumo} & \textbf{Avg.} \\
& \textit{pass@256} & \textit{pass@256} & \textit{pass@256} & \textit{pass@256} & \textit{pass@256} & \\
\midrule
\emph{Qwen2.5-Math-7B}   & 73.33 & 46.67 & 16.70 & 33.00 & 34.00 & 40.74 \\
\quad + DAPO             & 63.33 & 46.67 & \textbf{20.00} & \underline{35.00} & 46.70 & 42.34 \\
\quad + PPO              & 63.33 & 46.67 & 16.70 & 29.00 & 36.70 & 38.48 \\
\quad + Bernoulli 2-bin PPO & \underline{73.33} & \underline{53.33} & 16.67 & \underline{35.00} & \underline{50.00} & \underline{45.67} \\
\quad + One-hot PPO      & 66.67 & \textbf{56.67} & 16.70 & 31.00 & 46.70 & 43.55 \\
\quad + Two-hot PPO      & 70.00 & \textbf{56.67} & 13.30 & \underline{35.00} & \underline{50.00} & 45.00 \\
\rowcolor{blue!8}
\quad + HL-Gauss PPO (ours) & \textbf{76.67} & \underline{53.33} & \textbf{20.00} & \textbf{37.00} & \textbf{53.30} & \textbf{48.06} \\
\quad $\Delta$ &
\textcolor{ForestGreen}{+13.34} &
\textcolor{ForestGreen}{+6.66} &
\textcolor{ForestGreen}{+3.30} &
\textcolor{ForestGreen}{+8.00} &
\textcolor{ForestGreen}{+16.60} &
\textcolor{ForestGreen}{+9.58} \\
\bottomrule
\end{tabular}
}
\end{subtable}

\end{table*}

\paragraph{What the Bernoulli control isolates.}
The Bernoulli two-bin critic predicts $P(G{=}1\mid s)$ with a binary classification loss and is distinct from two-hot interpolation over the 101-bin support. It improves pass@256 over MSE (45.67 vs.\ 38.48) but does not improve avg@256 (15.50 vs.\ 15.88), and remains below HL-Gauss on both aggregates. Binary classification therefore helps recover rare successful trajectories, but does not fully account for the benefit of a smoothed, metric-aware value support.

\paragraph{Why pass@\emph{k} improves more.}
If a sample succeeds with probability $p$, then $\mathrm{pass}@k=1-(1-p)^k$ and its sensitivity is $k(1-p)^{k-1}$. Improvements on hard problems, where $p$ is small, can therefore be amplified in pass@\emph{k} even when their effect on avg@\emph{k} is modest. This is consistent with our calibration analysis: the largest critic gap occurs on likely-failure prefixes, where rare successful continuations are most vulnerable to under-reinforcement.

\subsection{Search-R1 benchmark}
\label{sec:search_r1_results}

To further evaluate reasoning with external search, we add experiments on Search-R1~\citep{jin2025searchr1} using the Qwen2.5-7B-Instruct~\citep{qwen2025qwen25technicalreport} backbone. We compare three training strategies under the same setup: DAPO, PPO, and HL-Gauss PPO. We report per-test-set metrics on seven datasets (NQ, TriviaQA, PopQA, HotpotQA, 2Wiki, MuSiQue, and Bamboogle) for transparent comparison.

\begin{table}[t]
\centering
\caption{Search-R1 results with mean@1 accuracy (\%). The upper block reports Search-R1-instruct numbers quoted from Search-R1~\citep{jin2025searchr1} for reference only. The lower block reports our runs on Qwen2.5-7B-Instruct, where \textbf{bold}/\underline{underlined} indicate best/second-best within this block. $\Delta$ denotes the performance gain of HL-Gauss PPO over PPO in percentage points.}
\label{tab:search-r1-results}
\setlength{\tabcolsep}{4pt}
\renewcommand{\arraystretch}{1.1}
\resizebox{\columnwidth}{!}{%
\begin{tabular}{lcccccccc}
\toprule
Method & NQ & TriviaQA & PopQA & HotpotQA & 2Wiki & MuSiQue & Bamboogle & Avg. \\
\midrule
\multicolumn{9}{l}{\emph{Reported in Search-R1~\citep{jin2025searchr1} (reference only)}} \\
Search-R1-instruct (GRPO) & 42.90 & 62.30 & 42.70 & 38.60 & 34.60 & 16.20 & 40.00 & 39.60 \\
Search-R1-instruct (PPO) & 39.30 & 61.00 & 39.70 & 37.00 & 41.40 & 14.60 & 36.80 & 38.50 \\
\midrule
Qwen2.5-7b-instruct & 32.13 & 53.32 & 25.37 & 27.63 & 7.50 & 28.00 & 40.00 & 30.56 \\
DAPO & 48.95 & \underline{65.54} & 42.23 & \underline{47.74} & 35.44 & 19.69 & \underline{45.60} & 43.60 \\
PPO & \underline{49.30} & 65.30 & \underline{44.40} & \underline{48.12} & \underline{41.00} & \underline{21.43} & 44.80 & \underline{44.91} \\
\rowcolor{blue!8}
HL-Gauss PPO (ours) & \textbf{50.86} & \textbf{67.29} & \textbf{46.26} & \textbf{49.64} & \textbf{43.38} & \textbf{22.06} & \textbf{50.40} & \textbf{47.13} \\
\quad $\Delta$ &
\textcolor{ForestGreen}{+1.56} &
\textcolor{ForestGreen}{+1.99} &
\textcolor{ForestGreen}{+1.86} &
\textcolor{ForestGreen}{+1.52} &
\textcolor{ForestGreen}{+2.38} &
\textcolor{ForestGreen}{+0.63} &
\textcolor{ForestGreen}{+5.60} &
\textcolor{ForestGreen}{+2.22} \\
\bottomrule
\end{tabular}
}
\end{table}

\begin{figure}[t]
\centering
\includegraphics[width=\linewidth]{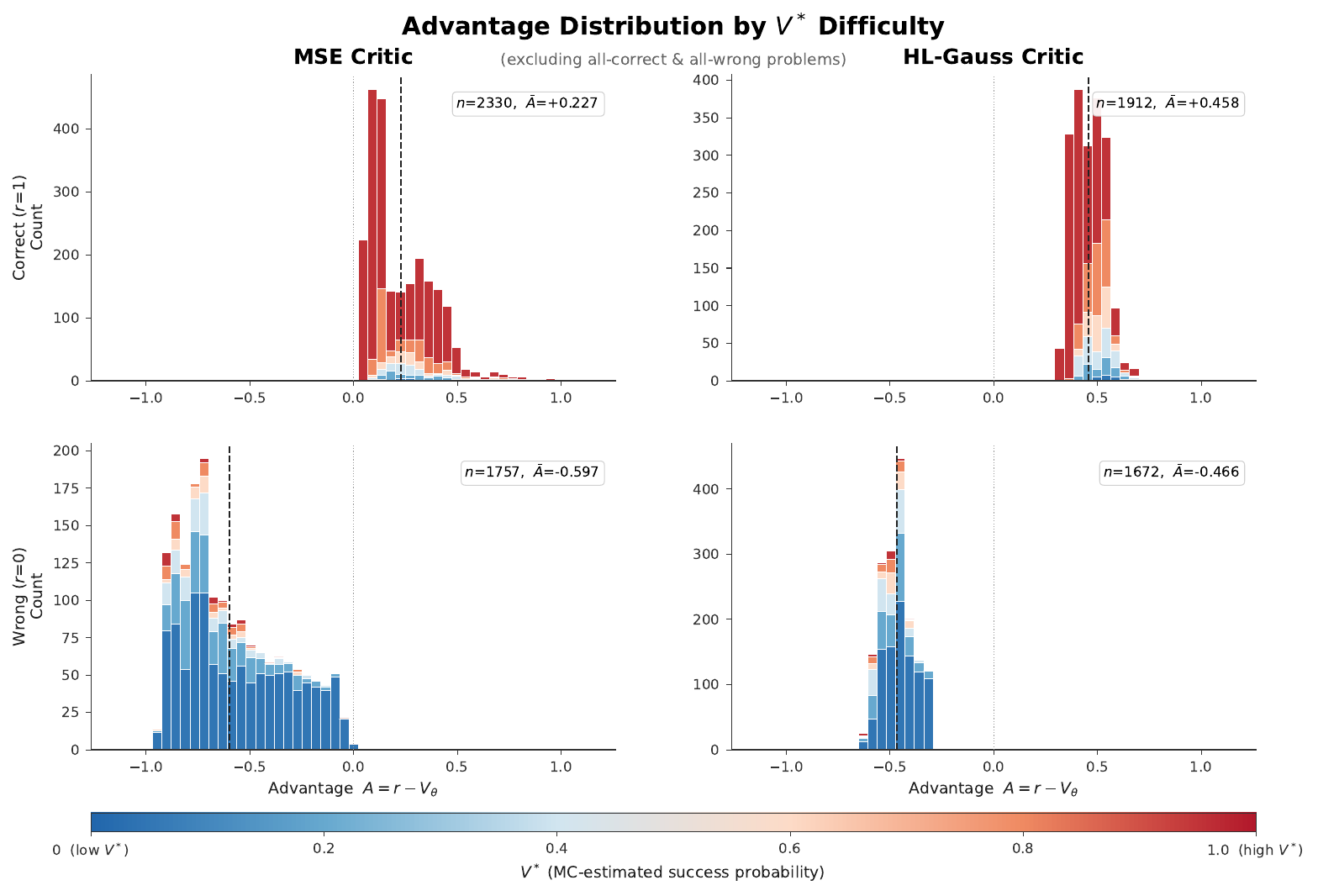}
\caption{Advantage distribution on AIME24 conditioned on episode outcome and oracle value $V^*$ (estimated via 256 Monte Carlo continuations). Each bar is colored by $V^*$, indicating the reasoning state's success probability. \textbf{Left:} MSE critic produces asymmetric advantages --- mean positive advantage ($+0.23$) is much smaller than mean negative advantage ($-0.60$). \textbf{Right:} HL-Gauss critic yields near-symmetric advantages (${\approx}{\pm}0.46$). All-correct and all-wrong problems are excluded.}
\label{fig:advantage-distribution}
\end{figure}

\begin{table}[t]
\centering
\setlength{\tabcolsep}{4pt}
\renewcommand{\arraystretch}{1.1}
\begin{minipage}{0.42\textwidth}
\small
\caption{Raw advantage statistics for MSE vs.\ HL-Gauss critics. Ratio near 1.0 indicates symmetric learning; higher values indicate the critic penalizes errors more than rewards successes.}
\label{tab:advantage-asymmetry}
\end{minipage}
\hfill
\begin{minipage}{0.55\textwidth}
\centering
\scriptsize
\begin{tabular}{llccc}
\toprule
Dataset & Critic & $\bar{A}_{r=0}$ & $\bar{A}_{r=1}$ & Ratio \\
\midrule
\multirow{2}{*}{AIME24} & MSE & $-0.597$ & $+0.227$ & 2.63$\times$ \\
 & HL-Gauss & $-0.466$ & $+0.458$ & 1.02$\times$ \\
\midrule
\multirow{2}{*}{AIME25} & MSE & $-0.486$ & $+0.175$ & 2.78$\times$ \\
 & HL-Gauss & $-0.450$ & $+0.456$ & 0.99$\times$ \\
\bottomrule
\end{tabular}
\end{minipage}
\end{table}

\subsection{Empirical analysis of advantage dynamics}
\label{sec:advantage_quality}

We now examine the \emph{advantage signals} produced by each critic, since these directly determine the PPO policy update. In our RLVR setting ($\gamma{=}1, \lambda{=}1$), with sparse binary rewards $r \in \{0,1\}$ assigned only at the terminal token, the GAE estimator reduces to a Monte Carlo residual: $\hat{A}_t = r - V_\phi(s_t)$. This makes advantage analysis a direct probe of critic calibration. We collect 64 rollouts per problem on AIME24 and AIME25 and evaluate critics from matched training protocols.

Table~\ref{tab:advantage-asymmetry} and Figure~\ref{fig:advantage-distribution} reveal a pronounced skew in the advantages produced by the MSE critic: penalties for incorrect rollouts ($r{=}0$) are 2--3$\times$ larger in magnitude than rewards for correct ones ($r{=}1$). On AIME24, for example, the mean raw advantage for failures is $-0.597$, whereas successful rollouts receive only $+0.227$ (ratio: $2.63\times$). The resulting policy update places more magnitude on failures than successes. In contrast, the HL-Gauss critic produces nearly symmetric advantages (ratio $\approx 1.0\times$). Batch-wise advantage normalization applies only an affine rescaling, $\tilde{A}_i=(A_i-\mu_B)/\sigma_B$; it standardizes scale but does not remove this relative asymmetry. The ratio is not expected to remain constant as the policy and critic evolve. Measurements at steps 30, 90, 150, and 300 show that MSE becomes increasingly negative-skewed while HL-Gauss remains closer to one (Appendix~\ref{app:advantage-over-training}).

\paragraph{Direct calibration.}
We additionally evaluate AIME24 prefixes at four positions per rollout, with oracle success probability $V^*(s)$ estimated from 256 Monte Carlo continuations. Table~\ref{tab:critic-calibration} reports the retained sample count and calibration metrics on this common prefix collection. Expectation decoding is the operational comparison because it is the scalar passed to GAE/PPO; mode decoding is included only as an auxiliary diagnostic. Relative to MSE, HL-Gauss expectation improves Brier score by 19.4\%, ECE by 18.4\%, and MCE by 27.9\%. The largest Brier gap occurs for $0<V^*<0.5$ (0.296 for MSE vs.\ 0.121 for HL-Gauss expectation), the region in which value overestimation most strongly inflates negative advantages. These post-hoc diagnostics support a calibration-based account, while not establishing advantage symmetry as the sole causal mediator of policy performance.

\begin{table}[t]
\centering
\small
\setlength{\tabcolsep}{4pt}
\renewcommand{\arraystretch}{1.05}
\caption{Critic calibration on AIME24 reasoning prefixes; lower is better. Oracle $V^*$ is estimated with 256 Monte Carlo continuations. ``Exp.'' is the scalar expectation used by PPO, while ``mode'' is auxiliary. Clamp is the fraction of predictions requiring projection to the valid binary range.}
\label{tab:critic-calibration}
\begin{tabular}{lrrrrr}
\toprule
Critic / decoder & $N$ & Brier & ECE & MCE & Clamp \\
\midrule
MSE & 7,671 & 0.2031 & 0.2612 & 0.4191 & 0.72\% \\
HL-Gauss exp. & 7,680 & 0.1637 & 0.2131 & 0.3021 & 0.00\% \\
HL-Gauss mode & 7,680 & \textbf{0.1470} & \textbf{0.1160} & \textbf{0.1624} & 0.00\% \\
\bottomrule
\end{tabular}
\end{table}

\section{Related work}
\label{sec:related}

\textbf{Reinforcement learning for LLMs.}
RLHF commonly uses PPO with learned reward models~\citep{ouyang2022training,stiennon2020learning,schulman2017ppo}. Critic-free alternatives include DPO~\citep{rafailov2023direct}, adaptive DPO sample scheduling~\citep{huang2025adaptive}, REINFORCE-style optimization~\citep{ahmadian2024back}, and GRPO~\citep{shao2024deepseekmath}. DPO optimizes preference pairs directly, REINFORCE uses Monte Carlo returns, and GRPO uses within-group reward comparisons. The success of DeepSeek-R1 has renewed interest in RLVR~\citep{deepseek2025r1}: DAPO adds dynamic sampling and clip-higher~\citep{yu2025dapo}, whereas VAPO improves actor--critic training through value pretraining and length-aware stabilization~\citep{zhong2025vapo}. Other work targets exploration and diversity using policy divergences~\citep{ICLR2026_6a1b224b}, count-based rewards~\citep{zhang2026count}, or asymmetric group-relative weighting~\citep{yu2026unveiling}. The last notion concerns symmetry of GRPO group weights, unlike the positive/negative scalar GAE balance analyzed here. These methods change data selection or policy optimization; we instead change only the critic learning objective, not rewards, policy regularization, or group weighting.

\textbf{Regression as classification for value functions.}
Discretizing continuous targets and training by classification has appeared throughout supervised learning and RL. Histogram losses provide smooth target distributions~\citep{imani2018improving}, MuZero uses two-hot categorical value and reward heads~\citep{schrittwieser2020mastering}, and broader studies show favorable scaling under non-stationary targets~\citep{farebrother2024stop}. These works establish the loss family; our contribution is its controlled study in PPO-based LLM RLVR. One-hot, two-hot, Bernoulli, smoothing, calibration, and advantage controls separate target construction from output dimensionality.

\textbf{Distributional reinforcement learning.}
Full-return methods include C51~\citep{bellemare2017distributional}, QR-DQN~\citep{dabney2018distributional}, D4PG~\citep{barthmaron2018d4pg}, nonlinear distributional gradient TD~\citep{qu2019nonlinear}, and LLM-oriented DisPPO~\citep{zhoudisppo}. They explicitly model $Z(s,a)$ and may exploit information beyond its mean for risk or exploration. Our categorical head instead serves only as a training objective and is decoded to the scalar $\hat{V}=\sum_i p_i z_i$ used by an unchanged GAE/PPO update; the actor never consumes quantiles, variance, or another higher-order statistic. It is therefore a drop-in replacement for the MSE critic rather than a full-return distributional method.

\section{Conclusion}
\label{sec:conclusion}

We study HL-Gauss as a drop-in critic-training surrogate for PPO in RLVR. The method replaces scalar MSE regression with smoothed categorical cross-entropy, then decodes a scalar expectation for an otherwise unchanged GAE/PPO actor update. Across mathematical reasoning, tool-augmented math, Search-R1, and Qwen2.5/Qwen3 backbones, this modification yields consistent gains over strong PPO and DAPO baselines. Bernoulli two-bin and matched categorical-head controls show that classification alone and output dimensionality do not fully explain the improvement. Direct calibration measurements further show that HL-Gauss produces better scalar value estimates on reasoning prefixes and more symmetric, lower-variance advantages. The evidence therefore supports a practical optimization-and-calibration benefit, rather than a claim that scalar expected-return critics are theoretically invalid.

\paragraph{Limitations.}
Our backbone coverage remains within the Qwen family and 4B--7B scale; broader architectures and larger models remain to be tested. The calibration and advantage analyses are diagnostic and do not prove that advantage symmetry is the only causal mediator of the policy gains. Finally, the best support margins and smoothing bandwidth can depend on the task's reward range, although the corresponding sensitivity and task-dependent settings are reported in the appendix.

\section{Ethics Statement}
This work proposes a methodological improvement to the critic training objective within PPO for LLMs. Our method does not introduce new data, models, or applications; it modifies an internal training component of existing RLVR pipelines. The benchmarks used are publicly available mathematical reasoning and open-domain QA datasets that do not contain sensitive or personally identifiable information. We do not foresee specific ethical risks arising from replacing MSE with classification-based value learning. Broader ethical considerations surrounding RLVR-trained LLMs --- such as the potential for generating plausible but incorrect reasoning --- apply equally to all methods in this space and are not unique to our contribution.

\section{Reproducibility statement}
We have made every effort to ensure that the results presented in this paper are reproducible. Base models and evaluation datasets are publicly available, and we will also release our checkpoints and datasets. Our code is openly available at: \url{https://github.com/ZhijianZhou/HL-guass-ppo}. The experimental setup, including training steps, training configurations, and hardware details, is described in detail in the appendix. We have also provided a full description to assist others in reproducing our experiments.

\paragraph{LLM usage disclosure.}
Large language models were used to assist with language polishing. All technical content, claims, and references were reviewed and verified by the authors.

\section*{Acknowledgments}
This work was supported by the National Natural Science Foundation of China (Grant Nos.~82394432 and 92249302) and the Shanghai Municipal Science and Technology Major Project (Grant No.~2023SHZDZX02).
\clearpage
\bibliography{colm2026_conference}
\bibliographystyle{colm2026_conference}

\appendix

\section{Detailed Training Configurations}
\label{app:detailed-configs}

\subsection{Training Data}

\paragraph{Mathematical Reasoning}
For both training and evaluation datasets, we use the following system prompt:
\begin{tcolorbox}[title=\textbf{System Prompt},colback=SeaGreen!10!CornflowerBlue!10,colframe=RoyalPurple!55!Aquamarine!100!]
Please reason step by step, and put your final answer within \textbackslash boxed\{\}.
\end{tcolorbox}

\paragraph{Math-with-tool}
For tool-augmented mathematical reasoning, we follow the standard chat template and use the built-in tool instruction prompt. Specifically, we prepend the following system prompt to each conversation in both training and evaluation:
\begin{tcolorbox}[
  title=\textbf{System Prompt},
  colback=SeaGreen!10!CornflowerBlue!10,
  colframe=RoyalPurple!55!Aquamarine!100!
]
You are Qwen, created by Alibaba Cloud. You are a helpful assistant.

\texttt{\# Tools}

You may call one or more functions to assist with the user query.

You are provided with function signatures within
\texttt{\textless tools\textgreater\textless/tools\textgreater}
XML tags:

\texttt{\textless tools\textgreater}

\texttt{\{"type": "function", "function": \{"name": "code\_interpreter", "description": "A tool for executing code.", "parameters": \{"type": "object", "properties": \{"code": \{"type": "string", "description": "The code to execute."\}\}, "required": ["code"]\}\}}

\texttt{\textless/tools\textgreater}

For each function call, return a JSON object with function name and arguments within
\texttt{\textless tool\_call\textgreater\textless/tool\_call\textgreater}
XML tags:

\texttt{\textless tool\_call\textgreater}

\texttt{\{"name": \textless function-name\textgreater, "arguments": \textless args-json-object\textgreater\}}

\texttt{\textless/tool\_call\textgreater}
\end{tcolorbox}

\paragraph{Search-R1}
For search-augmented reasoning, we follow the standard web information seeking paradigm. We prepend the following system prompt to each query in both training and evaluation:
\begin{tcolorbox}[
  title=\textbf{System Prompt},
  colback=SeaGreen!10!CornflowerBlue!10,
  colframe=RoyalPurple!55!Aquamarine!100!
]
You are a Web Information Seeking Master. Your task is to thoroughly seek the internet for information and provide accurate answers to questions. No matter how complex the query, you will not give up until you find the corresponding information.

As you proceed, adhere to the following principles:

\textbf{1. Persistent Actions for Answers}: You will engage in many interactions, delving deeply into the topic to explore all possible aspects until a satisfactory answer is found.

\textbf{2. Repeated Verification}: Before presenting a Final Answer, you will \textbf{cross-check} and \textbf{validate the information} you've gathered to confirm its accuracy and reliability.

\textbf{3. Attention to Detail}: You will carefully analyze each information source to ensure that all data is current, relevant, and from credible origins.
\end{tcolorbox}

\subsection{RL Training Configuration}

\paragraph{Mathematical Reasoning}

We use the hyperparameters in Table~\ref{tab:rl-math-config} for RL training on mathematical reasoning.

We use an outcome-based reward function that assigns $+1$ for correct final answers and $0$ otherwise.

\begin{table}[h]
\centering
\caption{RL training configurations for mathematical reasoning (Qwen2.5-Math-7B).}
\label{tab:rl-math-config}
\begin{tabular}{ll}
\toprule
Hyperparameter & Value \\
\midrule
Optimizer & AdamW \\
Actor learning rate & 1e-6 \\
Critic learning rate & 2e-6 \\
Training batch size & 512 \\
Samples per prompt & 16 \\
Mini-batch size & 32 \\
Max prompt length & 1024 \\
Max response length & 3072 \\
Rollout temperature & 1.0 \\
\bottomrule
\end{tabular}
\end{table}

\paragraph{Math-with-tool}

We use the hyperparameters in Table~\ref{tab:rl-math-tool-config} for RL training on tool-augmented mathematical reasoning.

For tool-augmented reasoning, we define an outcome-based reward with an auxiliary tool-usage term. Let $s$ denote the generated solution and $y$ the ground-truth answer. We first compute a base correctness score using strict boxed-answer verification:
\begin{equation}
r_{\text{ans}}(s, y) \in \{0, 1\},
\end{equation}
where $r_{\text{ans}}(s, y) = 1$ if the final boxed answer is correct and $0$ otherwise.

To account for tool usage, we define a lightweight turn-based bonus. Let $T$ denote the total number of interaction turns in the trajectory. The tool-related adjustment is given by
\begin{equation}
r_{\text{tool}}(T) = \frac{T - 2}{2} \cdot \alpha,
\end{equation}
where $\alpha = 0.1$ is a fixed scaling constant and the first two turns receive no bonus.

The final reward is computed as
\begin{equation}
r(s, y, T) =
\begin{cases}
r_{\text{ans}}(s, y) + \mathbb{I}\!\left[r_{\text{tool}}(T) > 0\right],
& \text{if } r_{\text{ans}}(s, y) = 1, \\[6pt]
\max\!\left(0,\; r_{\text{ans}}(s, y) + r_{\text{tool}}(T)\right),
& \text{otherwise},
\end{cases}
\end{equation}
where the auxiliary term is bounded so that incorrect final answers do not receive positive reward.

\begin{table}[h]
\centering
\caption{RL training configurations for tool-augmented mathematical reasoning (Qwen2.5-7B-Instruct).}
\label{tab:rl-math-tool-config}
\begin{tabular}{ll}
\toprule
Hyperparameter & Value \\
\midrule
Optimizer & AdamW \\
Actor learning rate & 1e-6 \\
Critic learning rate & 2e-6 \\
Training batch size & 128 \\
Samples per prompt & 16 \\
Mini-batch size & 32 \\
Max prompt length & 2048 \\
Max response length & 16384 \\
Max conversation turns & 8 \\
Rollout temperature & 1.0 \\
\bottomrule
\end{tabular}
\end{table}

\paragraph{Search-R1}

We use the hyperparameters in Table~\ref{tab:rl-search-r1-config} for RL training on Search-R1.

The reward function for Search-R1 serves as the primary training signal, guiding the optimization process in RL. We adopt a rule-based reward system that consists solely of final outcome rewards, which assess the correctness of the model's response. For factual reasoning tasks, correctness is evaluated using rule-based criteria such as exact string matching:
\begin{equation}
r_\phi(x, y) = \text{EM}(a_{\text{pred}}, a_{\text{gold}}),
\end{equation}
where $a_{\text{pred}}$ is the extracted final answer from response $y$ and $a_{\text{gold}}$ is the ground truth answer. The EM score returns $1$ for exact matches and $0$ otherwise.

\begin{table}[h]
\centering
\caption{RL training configurations for Search-R1 (Qwen2.5-7B-Instruct).}
\label{tab:rl-search-r1-config}
\begin{tabular}{ll}
\toprule
Hyperparameter & Value \\
\midrule
Optimizer & AdamW \\
Actor learning rate & 1e-6 \\
Critic learning rate & 2e-6 \\
Training batch size & 512 \\
Samples per prompt & 5 \\
Mini-batch size & 256 \\
Max prompt length & 2048 \\
Max response length & 16384 \\
Max conversation turns & 6 \\
Rollout temperature & 1.0 \\
\bottomrule
\end{tabular}
\end{table}

\subsection{Inference Configurations}

\paragraph{Mathematical Reasoning}
We use a rollout temperature of $0.6$, top-$p$ sampling with $p=0.95$, and a maximum response length of 4096 tokens. We adopt $k=256$ for AIME24/2025 datasets, $k=16$ for Minerva, MATH500, and OlympiadBench, and $k=8$ for additional benchmarks, balancing computational cost and task difficulty.

\paragraph{Math-with-tool}
We use a rollout temperature of $0.6$, top-$p$ sampling with $p=0.95$, and a maximum response length of 16384 tokens.

\paragraph{Search-R1}
We use a rollout temperature of $1.0$ during training and $0.0$ (greedy sampling) during evaluation, with top-$p$ sampling at $p=1.0$, and a maximum response length of 16384 tokens. We evaluate with mean@1 accuracy on seven datasets (NQ, TriviaQA, PopQA, HotpotQA, 2Wiki, MuSiQue, and Bamboogle).

\subsection{HL-Gauss Critic Configuration}

The HL-Gauss categorical critic extends the PPO critic with a categorical value head used as a training parameterization for a decoded scalar value. The following hyperparameters are specific to the HL-Gauss formulation:

\begin{table}[h]
\centering
\caption{HL-Gauss critic-specific configurations (task-dependent support ranges).}
\label{tab:hl-gauss-critic-config}
\begin{tabular}{lll}
\toprule
Hyperparameter & \multicolumn{2}{c}{Value} \\
\midrule
& Math Reasoning & Math-with-tool / Search-R1 \\
\midrule
Number of bins ($m$) & \multicolumn{2}{c}{101} \\
Support range $[v_{\min}, v_{\max}]$ & $[-0.1, 1.1]$ & $[-1.1, 2.2]$ / $[-1.1, 1.1]$ \\
Smoothing bandwidth ($\sigma$) & \multicolumn{2}{c}{0.009} \\
Critic warmup steps & \multicolumn{2}{c}{30} \\
Discount factor ($\gamma$) & \multicolumn{2}{c}{1.0} \\
GAE parameter ($\lambda$) & \multicolumn{2}{c}{1.0} \\
Gradient clipping & \multicolumn{2}{c}{1.0} \\
PPO clip range & \multicolumn{2}{c}{[0.2, 0.28]} \\
\bottomrule
\end{tabular}
\end{table}

The HL-Gauss binning scheme uses a uniform discretization of the value support into $m{=}101$ bins. The support range follows each task's attainable reward/value range: for binary mathematical reasoning we use $[-0.1,1.1]$; for tool-augmented reasoning with bonuses, $[-1.1,2.2]$; and for Search-R1 with possible negative rewards, $[-1.1,1.1]$. For binary math, using exactly $[0,1]$ would place both labels on support boundaries, truncating roughly half of the Gaussian mass before renormalization and making endpoint targets nearly one-hot. The 0.1 margin preserves smoothing at both endpoints. With $\sigma{=}0.009=0.75\Delta$, the target mass is concentrated over approximately 2--3 nearby bins, balancing sharpness and local smoothness.

\section{Additional experimental results}
\label{app:additional}

\subsection{Math-with-tool performance}
\label{app:math-with-tool}

Beyond pure mathematical reasoning on Qwen2.5-Math-7B, we also evaluate HL-Gauss PPO on tool-augmented mathematical reasoning using Qwen2.5-7B-Instruct. Under the same setup alignment, Table~\ref{tab:math-with-tool} reports tool-augmented math results. Compared with PPO, HL-Gauss PPO improves the average from 37.45 to 39.15 (+1.70), driven primarily by the AIME25 gain (+3.2), while remaining nearly unchanged on AIME24 (+0.2). This shows that the classification-based critic transfers effectively to tool-use reasoning scenarios under the same training and evaluation conditions.

\begin{table}[h]
\centering
\caption{Math-with-tool performance (Qwen2.5-7B-Instruct). Results reported as avg@64.}
\label{tab:math-with-tool}
\setlength{\tabcolsep}{6pt}
\renewcommand{\arraystretch}{1.1}
\begin{tabular}{lccc}
\toprule
\quad Methods & \textbf{AIME24} & \textbf{AIME25} & \textbf{Avg.} \\
& \textit{avg@64} & \textit{avg@64} & \\
\midrule
Qwen2.5-7b-instruct & 12.7 & 5.5& 9.10 \\
\quad + DAPO             & 37.4 & 29.3 & 33.35 \\
\quad + PPO              & \underline{43.9} & \underline{31.0} & \underline{37.45} \\
\rowcolor{blue!8}
\quad + HL-Gauss PPO (ours) & \textbf{44.1} & \textbf{34.2} & \textbf{39.15} \\
\quad $\Delta$ (vs PPO) &
\textcolor{ForestGreen}{$+$0.2} &
\textcolor{ForestGreen}{+3.2} &
\textcolor{ForestGreen}{+1.7} \\
\bottomrule
\end{tabular}
\end{table}

\subsection{Generalization to Qwen3-4B-Base}
\label{app:qwen3-generalization}

To test whether the main result is confined to the Qwen2.5-7B backbone, we repeat the matched PPO comparison with Qwen3-4B-Base. Table~\ref{tab:qwen3-generalization} shows improvements on all five benchmarks. The aggregate gains are +1.85 avg@256 and +3.73 pass@256.

\begin{table}[h]
\centering
\caption{Matched PPO results on Qwen3-4B-Base. HL-Gauss changes only the critic head and loss.}
\label{tab:qwen3-generalization}
\setlength{\tabcolsep}{4pt}
\renewcommand{\arraystretch}{1.05}
\resizebox{\textwidth}{!}{%
\begin{tabular}{llcccccc}
\toprule
Metric & Critic & AIME24 & AIME25 & HMMT & BeyondAIME & Brumo & Avg. \\
\midrule
\multirow{2}{*}{avg@256}
& MSE & 21.91 & 21.31 & 7.98 & 9.50 & 15.96 & 15.33 \\
& HL-Gauss & \textbf{25.78} & \textbf{22.97} & \textbf{8.76} & \textbf{10.93} & \textbf{17.48} & \textbf{17.18} \\
\midrule
\multirow{2}{*}{pass@256}
& MSE & 67.70 & 63.30 & 33.33 & 46.00 & 56.67 & 53.40 \\
& HL-Gauss & \textbf{70.00} & \textbf{70.00} & \textbf{36.67} & \textbf{49.00} & \textbf{60.00} & \textbf{57.13} \\
\bottomrule
\end{tabular}
}
\end{table}

\subsection{Ablation: Smoothing bandwidth (\texorpdfstring{$\sigma$}{sigma})}
\label{app:ablation-sigma}

We evaluate the sensitivity of HL-Gauss PPO to the smoothing bandwidth parameter $\sigma$, which controls the width of the Gaussian kernel applied to target value distributions. We test three values on mathematical reasoning benchmarks (Qwen2.5-Math-7B): $\sigma = 0.009$ (our default, $0.75\Delta$), $\sigma = 0.012$ ($\Delta$), and $\sigma = 0.024$ ($2\Delta$), where $\Delta \approx 0.012$ is the bin width.

\begin{table}[h]
\centering
\caption{Performance of HL-Gauss PPO under different smoothing bandwidth ($\sigma$) values on mathematical reasoning. Best results are shown in \textbf{bold} and second-best results are \underline{underlined}.}
\label{tab:ablation-sigma}
\setlength{\tabcolsep}{4pt}
\renewcommand{\arraystretch}{1.1}

\begin{subtable}{\textwidth}
\centering
\caption{avg@256 results}
\begin{tabular}{lcccccc}
\toprule
Smoothing & \textbf{AIME24} & \textbf{AIME25} & \textbf{HMMT} & \textbf{BeyondAIME} & \textbf{Brumo} & \textbf{Avg.} \\
$\sigma$ & avg@256 & avg@256 & avg@256 & avg@256 & avg@256 & \\
\midrule
$\sigma = 0.024$ (2$\Delta$) &34.90 &18.06 &3.56  &7.40  &18.50  &16.48 \\
$\sigma = 0.012$ ($\Delta$) & \underline{35.05} &\underline{18.90} &\underline{4.53}  &\underline{7.50}  &\underline{19.00}  &\underline{17.00} \\
$\sigma = 0.009$ (0.75$\Delta$)  & \textbf{35.20} & \textbf{23.80} & \textbf{4.78} & \textbf{8.50} & \textbf{21.40} & \textbf{18.74} \\
\bottomrule
\end{tabular}
\end{subtable}

\vspace{0.8em}

\begin{subtable}{\textwidth}
\centering
\caption{pass@256 results}
\begin{tabular}{lcccccc}
\toprule
Smoothing & \textbf{AIME24} & \textbf{AIME25} & \textbf{HMMT} & \textbf{BeyondAIME} & \textbf{Brumo} & \textbf{Avg.} \\
$\sigma$ & pass@256 & pass@256 & pass@256 & pass@256 & pass@256 & \\
\midrule
$\sigma = 0.024$ (2$\Delta$) &\underline{73.33}  &\underline{53.33}  &\underline{16.70}  &34.10  &45.00  &44.49 \\
$\sigma = 0.012$ ($\Delta$) &66.67  &\textbf{60.00}  &\textbf{20.00}  &\underline{35.00}  &\underline{48.30}  &\underline{45.99} \\
$\sigma = 0.009$ (0.75$\Delta$)  & \textbf{76.67} & \underline{53.33} & \textbf{20.00} & \textbf{37.00} & \textbf{53.30} & \textbf{48.06} \\
\bottomrule
\end{tabular}
\end{subtable}

\end{table}

\textbf{Analysis.}
The results clearly demonstrate that smaller smoothing bandwidth values yield better performance across all benchmarks. With $\sigma = 0.009$ (0.75$\Delta$), HL-Gauss PPO achieves avg@256 of 18.74 and pass@256 of 48.06, compared to 17.00 and 45.99 for $\sigma = 0.012$ ($\Delta$), and 16.48 and 44.49 for $\sigma = 0.024$ (2$\Delta$).

This trend is consistent across individual datasets (AIME24, AIME25, HMMT, BeyondAIME, Brumo), suggesting that tighter smoothing is preferable for the RLVR regime. Intuitively, smaller $\sigma$ provides more discriminative target distributions that better distinguish between value regions, while larger $\sigma$ over-smooths and blurs fine-grained value distinctions. Our choice of $\sigma = 0.75\Delta$ thus represents an effective balance, placing sufficient probability mass at the target value while maintaining sharpness in the surrounding bins. These results validate the default configuration used throughout the main experiments and suggest that further tuning toward even smaller $\sigma$ may yield diminishing returns due to numerical stability considerations.

\subsection{Advantage error decomposition by V* region}
\label{app:advantage-error-region}

\begin{figure}[h]
\centering
\includegraphics[width=\linewidth]{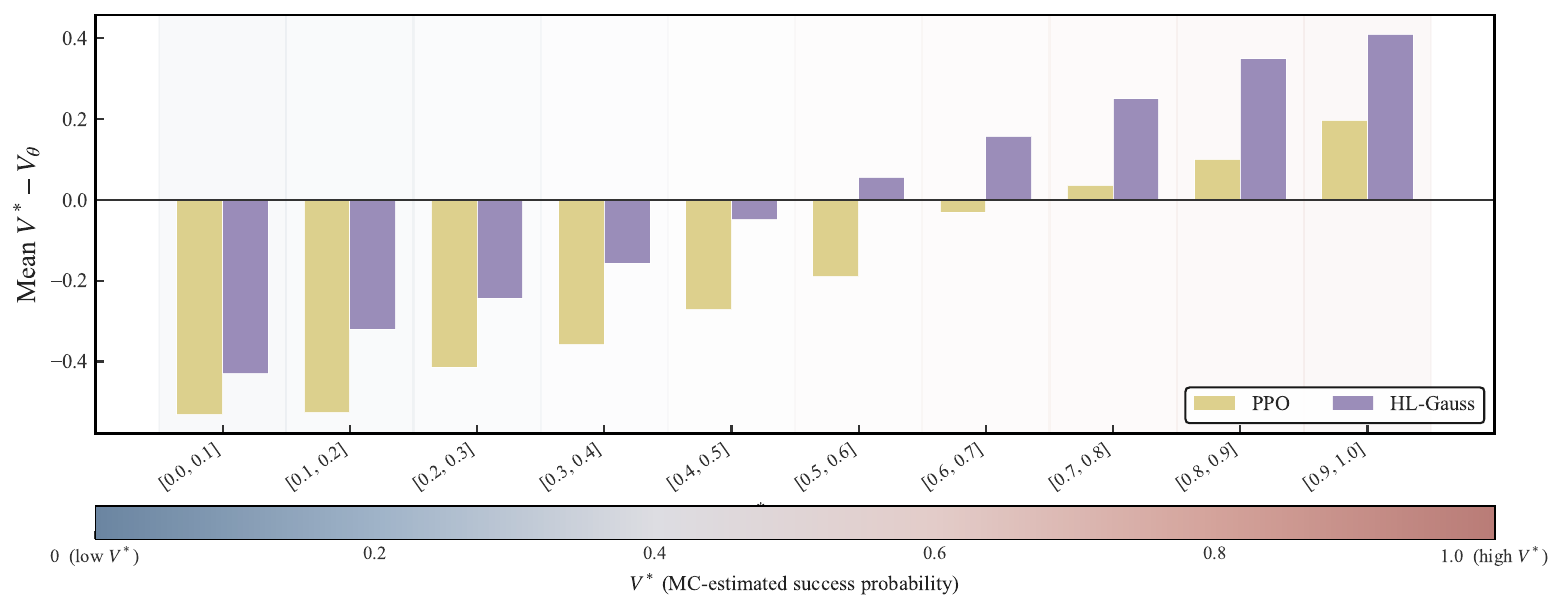}
\caption{Signed advantage bias ($V^* - V_\phi$) decomposed by oracle value V* region (10 bins, excluding all-correct and all-wrong problems). PPO (MSE critic) shows strong negative bias in low-V* regions (systematic value overestimation) and positive bias in high-V* regions (underestimation), while HL-Gauss critic maintains near-zero bias across all V* regions, indicating superior value calibration.}
\label{fig:advantage-error-region}
\end{figure}

To further isolate the bias in critic predictions across different oracle value regions, we stratify advantages by the oracle value $V^*(s_t)$ (estimated via 256 Monte Carlo continuations). Figure~\ref{fig:advantage-error-region} shows the signed bias ($V^* - V_\phi$) across ten V* bins. The MSE critic exhibits a systematic shift: it overestimates values in the likely-wrong region ($V^*{\approx}0$) and underestimates them in the likely-correct region ($V^*{\approx}1$).

This stratification localizes the empirical calibration gap: the MSE critic overestimates likely-failure prefixes and therefore suppresses the positive advantage assigned to rare successful continuations. HL-Gauss predictions stay closer to the oracle estimates across these regions, consistent with the aggregate Brier/ECE results in Table~\ref{tab:critic-calibration}.

\subsection{Advantage asymmetry over training}
\label{app:advantage-over-training}

The negative-to-positive advantage ratio evolves as the policy distribution and critic calibration change, so it need not be constant. Table~\ref{tab:advantage-over-training} reports AIME24 measurements across training. The MSE ratio grows from 0.134 to 2.63, whereas HL-Gauss remains closer to one at the final checkpoint. This temporal pattern shows that the main-text result is not selected from an isolated fluctuation.

\begin{table}[h]
\centering
\caption{AIME24 raw advantage means over training. Ratio is $|\bar A_{\mathrm{wrong}}|/\bar A_{\mathrm{correct}}$; values near one indicate balanced magnitudes.}
\label{tab:advantage-over-training}
\setlength{\tabcolsep}{4pt}
\renewcommand{\arraystretch}{1.05}
\begin{tabular}{rrrrrrr}
\toprule
& \multicolumn{3}{c}{MSE} & \multicolumn{3}{c}{HL-Gauss} \\
\cmidrule(lr){2-4}\cmidrule(lr){5-7}
Step & $\bar A_c$ & $\bar A_w$ & Ratio & $\bar A_c$ & $\bar A_w$ & Ratio \\
\midrule
30  & +0.777 & -0.104 & 0.134 & +0.670 & -0.226 & 0.337 \\
90  & +0.512 & -0.404 & 0.790 & +0.527 & -0.360 & 0.684 \\
150 & +0.312 & -0.538 & 1.724 & +0.472 & -0.403 & 0.854 \\
300 & +0.227 & -0.597 & 2.630 & +0.458 & -0.466 & 1.020 \\
\bottomrule
\end{tabular}
\end{table}

\section{Computational Overhead}
\label{app:compute}

The HL-Gauss categorical critic introduces minimal parameter overhead compared to the standard MSE critic. The only architectural difference is the value head output dimension: $\mathbb{R}^d \to \mathbb{R}^{101}$ (categorical) versus $\mathbb{R}^d \to \mathbb{R}^1$ (scalar), where $d$ is the hidden dimension. This increases the value head parameters by a factor of 101, but since the value head constitutes a negligible fraction of the LLM backbone ($\sim 0.006\%$ for Qwen2.5-7B), the overall model size increase is imperceptible.

In terms of computational cost per training step, the categorical critic adds:
\begin{itemize}
    \item Softmax computation over 101 bins (negligible)
    \item Cross-entropy loss instead of MSE (negligible difference)
\end{itemize}
These overheads are dwarfed by the LLM forward/backward pass, resulting in no meaningful increase in wall-clock training time. Empirically, HL-Gauss PPO and standard PPO exhibit near-identical per-step training speed on the same hardware, confirming that the architectural change is computationally efficient.

\begin{figure}[h]
\centering
\includegraphics[width=\linewidth]{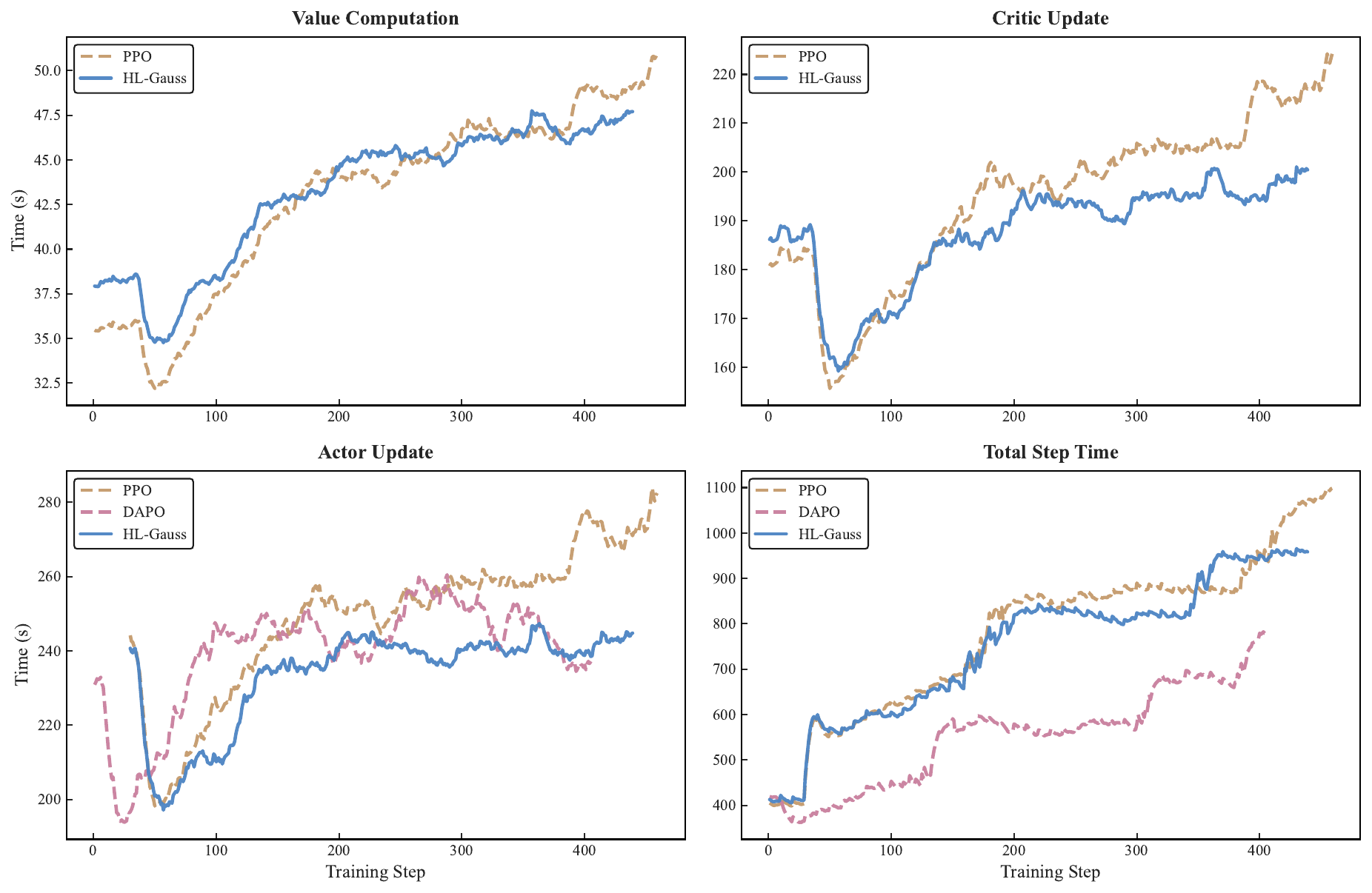}
\caption{Training step timing breakdown across 400 steps. Value Computation and Critic Update show negligible differences between PPO and HL-Gauss. Actor Update and Total Step Time are nearly identical, confirming minimal computational overhead from the categorical head. }
\label{fig:timing-breakdown}
\end{figure}

\section{Method Implementation Details}
\label{app:method-impl}

\textbf{Support range and resolution.}
In RLVR with binary rewards and $\gamma{=}1$, the theoretical value range is $[0,1]$. We set $v_{\min}{=}{-0.1}$ and $v_{\max}{=}1.1$ to provide a small margin beyond the reward boundaries, and use $m{=}101$ bins (bin width $\Delta{\approx}0.0119$). This resolution is sufficient to represent fine-grained value distinctions while adding negligible computational overhead (Appendix ~\ref{app:compute}).

\textbf{Smoothing bandwidth.}
We set $\sigma{=}0.009 = 0.75\Delta$, which provides moderate smoothing across approximately 2--3 bins around each target. This balances between the extremes of one-hot encoding ($\sigma{\to}0$, no smoothing) and overly diffuse targets ($\sigma{\gg}\Delta$, which blur value distinctions).

\textbf{Critic warmup.}
Following prior work on PPO training for LLMs~\citep{yu2025dapo}, we warm up the critic for 30 steps before beginning actor updates. During warmup, only the critic is trained on rollout data from the frozen initial policy, allowing the value estimates to stabilize before they influence advantage computation.

\textbf{Architecture.}
The critic is initialized from the same pretrained backbone as the actor. The only architectural difference from the standard PPO critic is the output head: $\mathbb{R}^d \to \mathbb{R}^m$ (categorical) instead of $\mathbb{R}^d \to \mathbb{R}^1$ (scalar). Actor and critic do not share backbone parameters.

\clearpage
\section{Case Study: Paired Correct and Wrong Rollouts}
\label{app:case-study}

We conclude with a token-level case study on a single AIME-style prompt for which our sampler produced both a correct rollout and a wrong rollout. The underlying math is straightforward:
\[
17_b = b + 7,
\qquad
97_b = 9b + 7.
\]
Therefore,
\[
(b + 7) \mid (9b + 7)
\iff
(b + 7) \mid \bigl(9(b + 7) - (9b + 7)\bigr)
\iff
(b + 7) \mid 56.
\]
Since $b > 9$, we have $b + 7 > 16$, so the only positive divisors of $56$ that remain possible are $28$ and $56$. Hence $b \in \{21, 49\}$, and the required sum is $70$. Thus, the rollout ending in $70$ is correct, while the rollout ending in $49$ is incorrect.

The wrong trace fails only after a locally plausible start. Its base conversion and divisibility setup are correct, but after introducing an auxiliary variable $k$ it effectively narrows $9-k$ to $\{\pm 1, \pm 7\}$, which is too restrictive and drops the valid case $b = 21$. By contrast, the correct trace keeps the simpler divisor argument $(b + 7) \mid 56$ and directly recovers both valid bases.

The heatmaps below show how the HL-Gauss critic tracks this divergence at the token level. In the correct rollout, predicted values remain high throughout the clean divisor-based derivation and stay high through the final answer $70$. In the wrong rollout, the critic assigns moderate value while the setup is still plausible, but its score collapses once the reasoning commits to the incomplete case split and remains low through the final answer $49$. This example illustrates that the categorical critic does not merely reward fluent math text; it distinguishes trajectories that stay on the correct proof path from trajectories that only appear locally reasonable.

\begin{figure}[p]
\centering
\includegraphics[page=1,width=\linewidth]{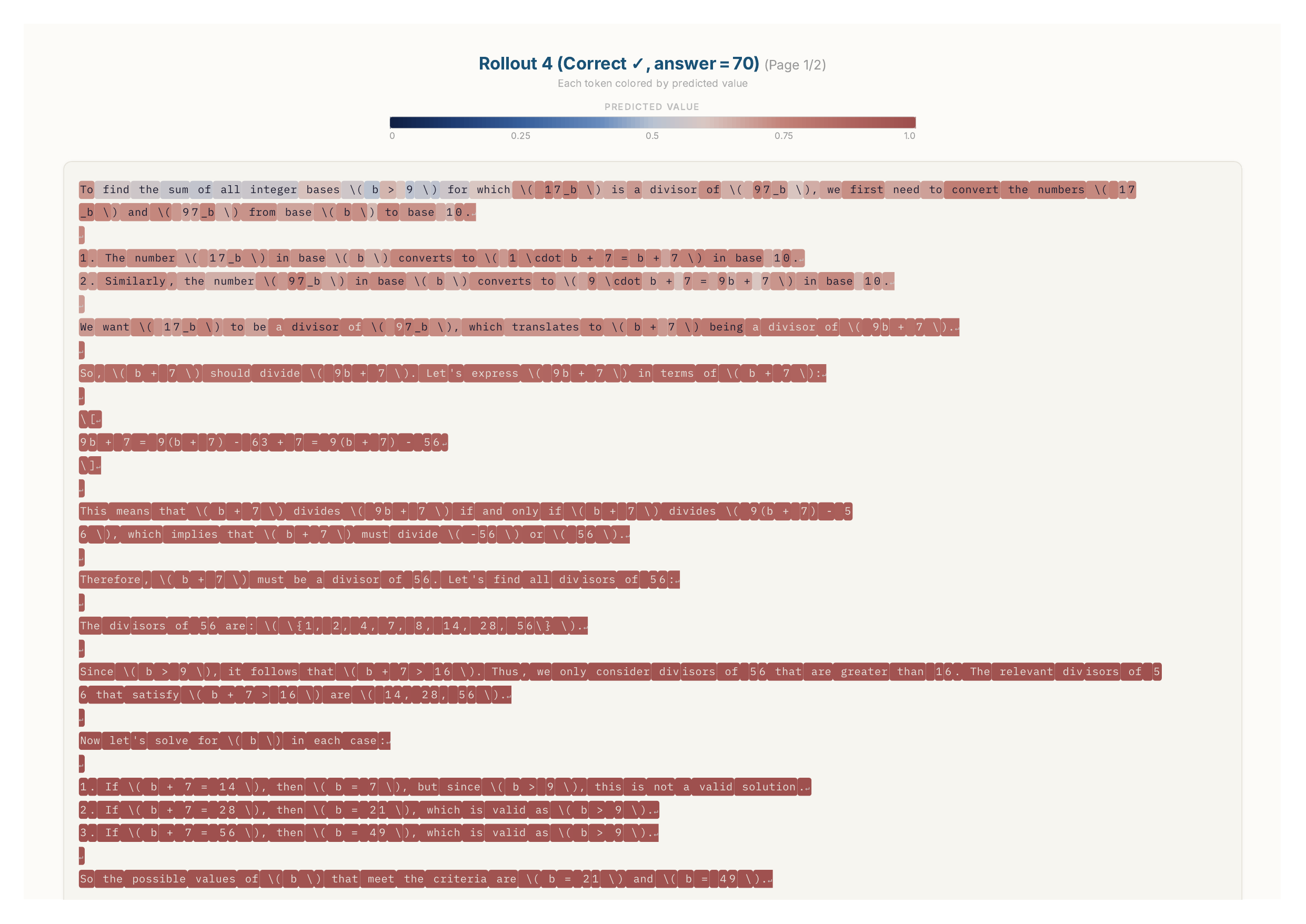}
\caption{Correct rollout for the paired case study (page 1/2). The trace quickly reduces the problem to the key divisor condition $(b + 7) \mid 56$, and the critic assigns consistently high value throughout this derivation.}
\label{fig:case-study-correct-p1}
\end{figure}

\begin{figure}[p]
\centering
\includegraphics[page=1,width=\linewidth]{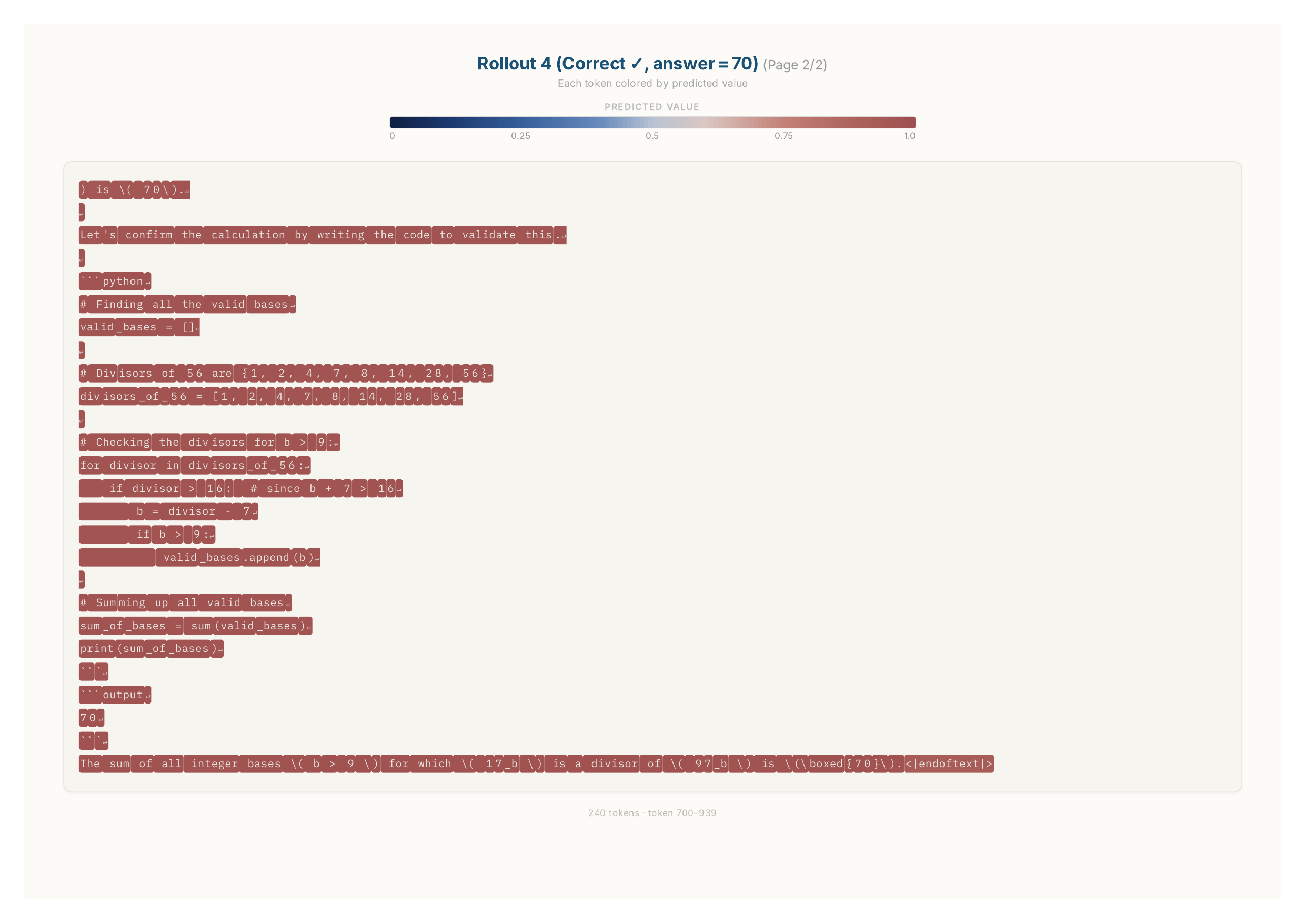}
\caption{Correct rollout for the paired case study (page 2/2). The critic remains confident through the identification of $b = 21$ and $b = 49$, and stays high through the final correct answer $70$.}
\label{fig:case-study-correct-p2}
\end{figure}

\begin{figure}[p]
\centering
\includegraphics[page=1,width=\linewidth]{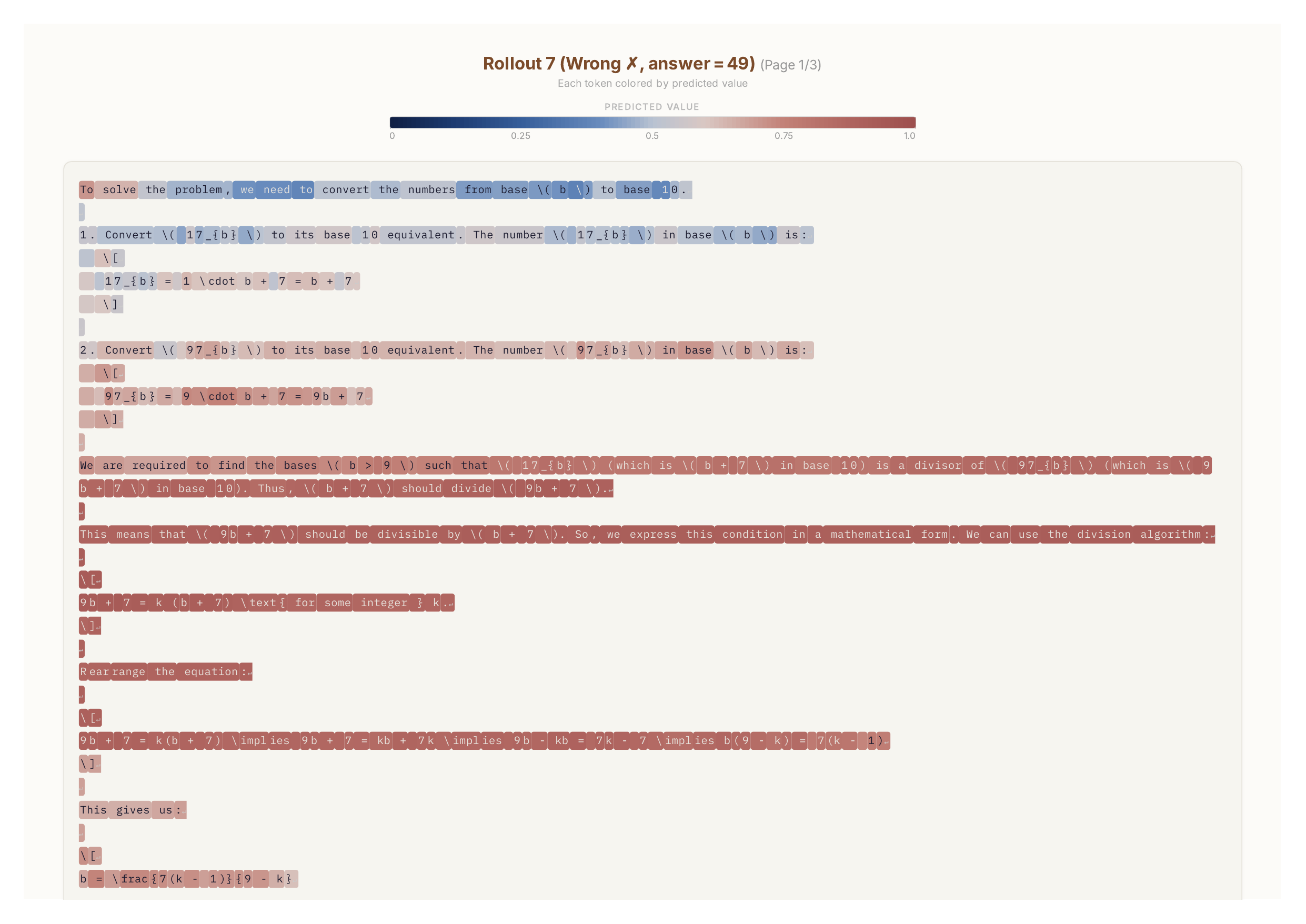}
\caption{Wrong rollout for the paired case study (page 1/3). The early setup is partly valid, so the critic does not immediately reject the trace even though the rollout is already drifting toward an overly restrictive analysis.}
\label{fig:case-study-wrong-p1}
\end{figure}

\begin{figure}[p]
\centering
\includegraphics[page=1,width=\linewidth]{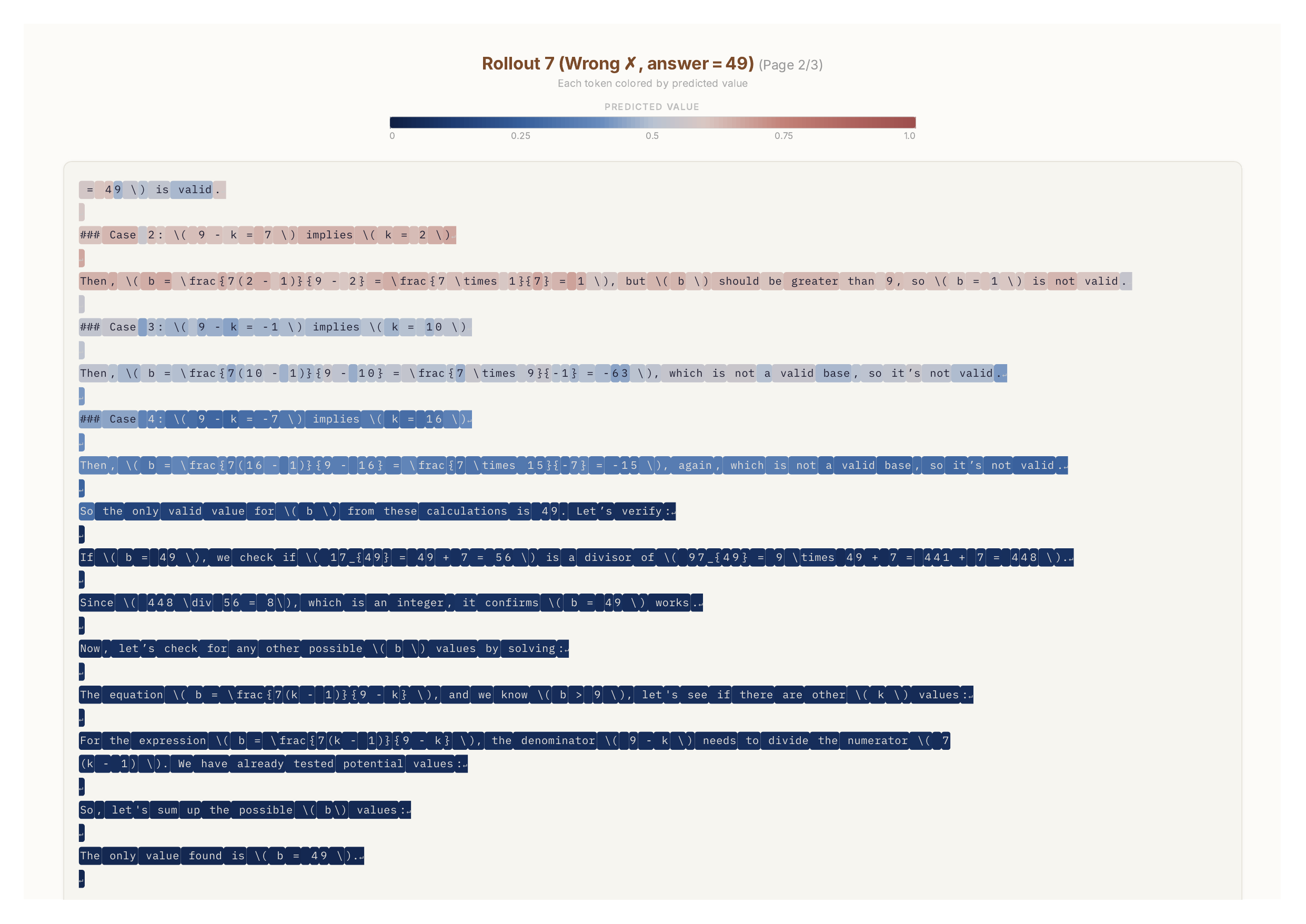}
\caption{Wrong rollout for the paired case study (page 2/3). Once the derivation commits to the incomplete $k$-based case split and excludes the valid base $b = 21$, the predicted value drops sharply.}
\label{fig:case-study-wrong-p2}
\end{figure}

\begin{figure}[p]
\centering
\includegraphics[page=1,width=\linewidth]{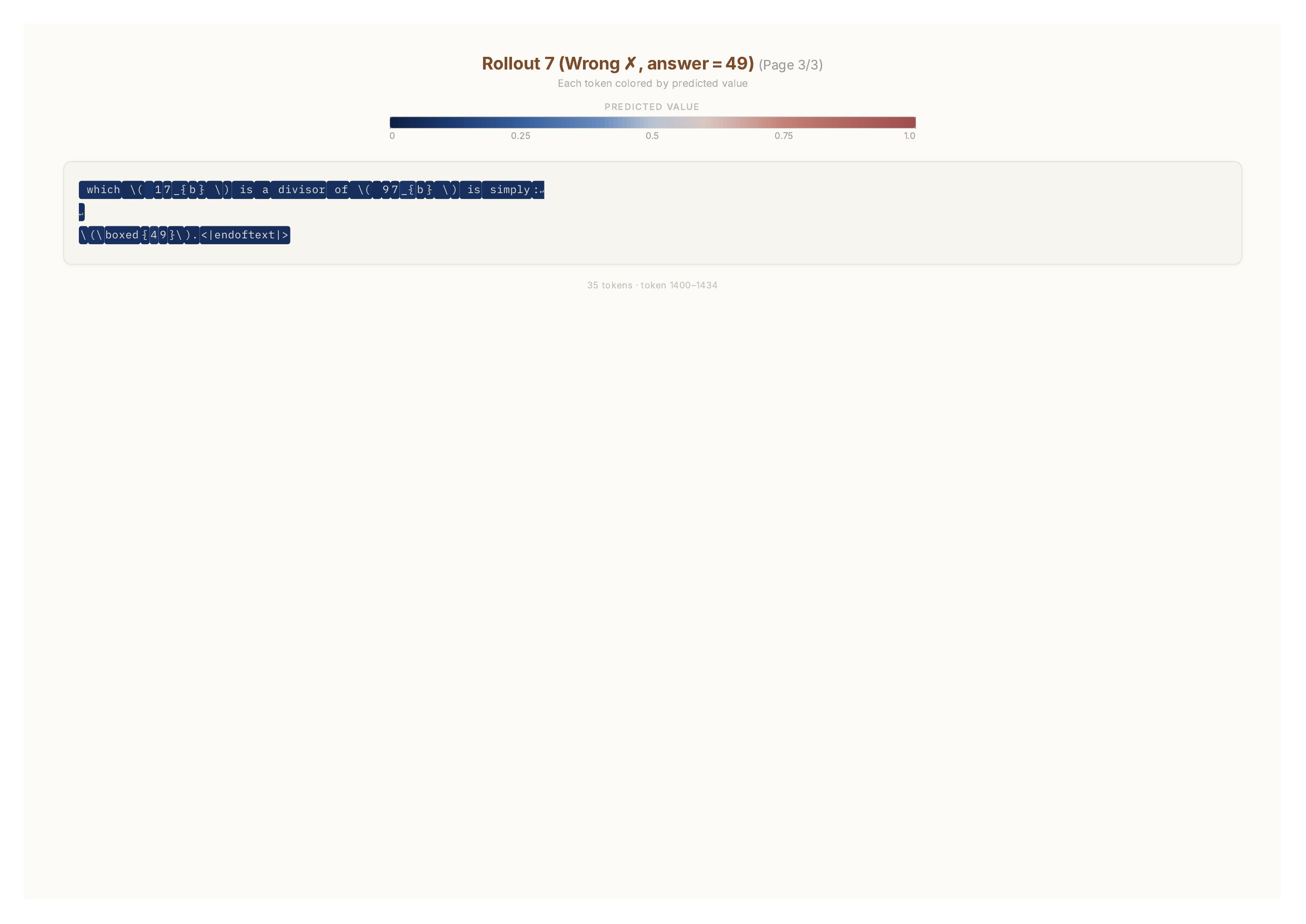}
\caption{Wrong rollout for the paired case study (page 3/3). The critic remains low through the final incorrect answer $49$, showing that it can separate a fluent-looking but invalid continuation from the correct solution path.}
\label{fig:case-study-wrong-p3}
\end{figure}

\clearpage
\end{document}